# Managing Requirement Volatility in an Ontology-Driven Clinical LIMS Using Category Theory

(International Journal of Telemedicine and Applications)

ARASH SHABAN-NEJAD[1,§], OLGA ORMANDJIEVA[1], MOHAMAD KASSAB[1], VOLKER HAARSLEV[1]

[1]Department of Computer Science and Software Engineering, Concordia University, 1455 de Maisonneuve Blvd. W., Montreal, Quebec H3G 1M8, Canada

E-mail: {arash_sh, ormandj, moh_kass, haarslev}@cs.concordia.ca

[§] Corresponding author

**Abstract**

Requirement volatility is an issue in software engineering in general, and in Web-based clinical applications in particular, which often originates from an incomplete knowledge of the domain of interest. With advances in the health science, many features and functionalities need to be added to, or removed from, existing software applications in the biomedical domain. At the same time, the increasing complexity of biomedical systems makes them more difficult to understand, and consequently it is more difficult to define their requirements, which contributes considerably to their volatility. In this paper, we present a novel agent-based approach for analyzing and managing volatile and dynamic requirements in an ontology-driven laboratory information management system (LIMS) designed for Web-based case reporting in medical mycology. The proposed framework is empowered with ontologies and formalized using category theory to provide a deep and common understanding of the functional and nonfunctional requirement hierarchies and their interrelations, and to trace the effects of a change on the conceptual framework.

## 1. Introduction

The life sciences constitute a challenging domain in knowledge representation. Biological data are highly dynamic, and bioinformatics applications are large and there are complex interrelationships between their elements with various levels of interpretation for each concept. In an ideal situation, the requirements for a software system should be completely and unambiguously determined before design, coding, and testing take place. The complexity of bioinformatics applications and their constant evolution lead to frequent changes in their requirements: often new requirements are added and existing requirements are modified or deleted, causing parts of the software system to be redesigned, deleted, or added. Such changes lead to volatility in the requirements of bioinformatics applications.

In this paper, we deal with an important problem of requirements volatility in the context of an ontology-driven clinical laboratory information management system (LIMS)[1, 2]. A LIMS is a software application for managing information about laboratory samples, users, instruments, standards, and other laboratory functions and products. It forms an essential part of electronic laboratory reporting (ELR) and electronic communicable disease reporting (CDR). ELR is a key factor in public health surveillance, improving real-time decision making based on messages reporting cases of notifiable conditions from multiple laboratories [3].Combining these reports with clinical experiments and case studies makes up a CDR system [4]. This framework, along with the active participation of physicians specializing in fungal infectious diseases, infection control professionals, and lab technicians, aimed at generating automated online reporting from clinical laboratories to improve the quality of lab administration, health surveillance, and disease notification. It provides security, portability, and accessibility over the Web, as well as efficiency and data integrity in clinical, pharmaceutical, industrial, and environmental laboratory processes.

**Research Problem:** Requirements volatility is "a measure of how much program requirements change once coding begins" [5]. Bioinformatics applications with frequently changing requirements have a high degree of volatility, while projects with relatively stable requirements have a low one [6]. Higher requirement volatility will result in higher development and maintenance costs, the risk of schedule slippage, and an overall decrease in the quality of the services provided. Therefore, requirement volatility is considered one of the major obstacles to using a LIMS. In this paper, we propose an innovative approach for the automatic tracing of volatile requirement changes based on their formal representation in an ontological framework using a solid mathematical foundation, namely, category theory [7].

**Approach:** Investigating the factors that drive requirement change is an important prerequisite for understanding the nature of requirement volatility. This increased understanding will minimize that volatility and improve the process of requirement change management. One of the most important volatility factors is the diversity of requirement definitions in the application domain, which may lead to confusing and frustrating communication problems between application users and software engineers [8]. Ontologies [9] are widely used as a vehicle for knowledge management sharing common vocabularies, describing the semantics of programming interfaces, providing a structure to organize knowledge, reducing the development effort for generic tools and systems, improving data and tool integration, reusing organizational knowledge, and capturing behavioral knowledge. Ontologies can describe software architectures and requirements, which are difficult to model with object oriented languages [10]. Conceptualization of the requirements using an ontology formalized with category theory minimizes requirement volatility by providing a deep and common understanding of the requirements [11], which is essential in order for bioinformatics application developers to manage the changes successfully. This paper proposes a generic categorical model of LIMS requirements with an emphasis on nonfunctional requirements, their dependencies and interdependencies using category theory as an advanced mathematical formalism. The resulting categorical model represents the functional requirements (FRs) and nonfunctional requirements (NFRs) based on an investigation of their dependencies and interdependencies, which is considered critical to success in tracing requirement changes. Requirement traceability, defined as "the ability to describe and follow the life of a requirement in both [forward and backward directions]" [12], is an essential part in performing requirement maintenance and change management processes. Moreover, the extent to which change traceability is exploited is viewed as an indicator of system quality and process maturity, and is mandated by existing standards [13]. These changes have to be monitored for consistency with the existing categorical framework in the LIMS context. After capturing the LIMS requirements in an ontological framework—to provide a common shared understanding of the requirements—empowered with category theory, a novel agent-based framework for the representation, legitimation, and reproduction (RLR) of changes [14] is proposed for implementing volatile requirement identification, and integrated change management and consistency monitoring in a LIMS (Figure 1).

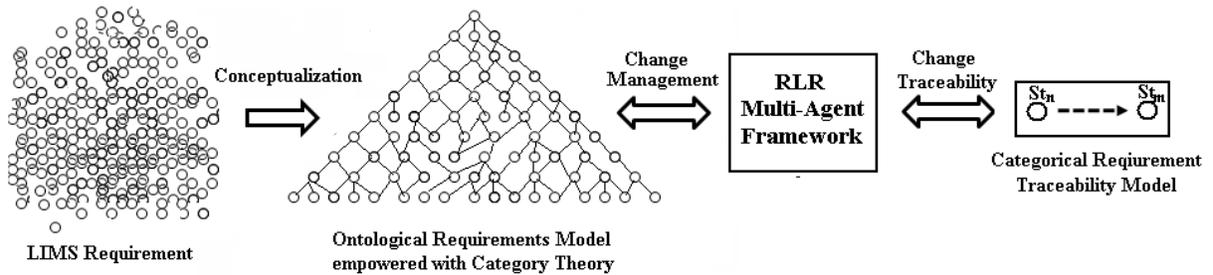

**Figure 1:** General view on the proposed approach for managing requirement volatility.

RLR framework assists and guides the software developer through the change management process in general, and in representing and tracing the changes, particularly through the use of category theory.

The rest of the paper is organized as follows. Our discussion will be illustrated through examples from the LIMS system case study introduced in Section 2. Our approach for recruiting category theory for formalizing the conceptual framework of the requirements is presented in Section 3. The RLR framework for managing changes is described in Section 4. In Sections 5 and 6, we demonstrate the applicability of our categorical method for representing and tracking requirement changes and formalizing the interaction of agents in the RLR framework through an application scenario. We describe the evaluation phase in the proposed multiagent framework and review related work in Sections 7 and 8, respectively.

The paper concludes with the list of contributions and an outline of research directions in Section 9.

## 2. The MYCO-LIMS Requirements Overview

The mycology laboratory information management system (MYCO-LIMS) is software for managing information about laboratory samples, users, instruments, standards, and other laboratory functions and products, and provides security, portability, and accessibility over the Web, efficiency, and data integrity in clinical, pharmaceutical, and industrial laboratory processes. The MYCO-LIMS is an ontology-driven object-oriented application for a typical fungal genomics lab performing sequencing and gene expression experiments in the domain of medical mycology. Based on Gruber's definition [9], an ontology is a "specification of conceptualization", and provides an underlying discipline for knowledge sharing by defining concepts, properties, and axioms. The term "conceptualization" includes conceptual frameworks for analyzing shared domain knowledge which are necessary for knowledge representation in the domain of interest. In our context, the conceptual framework for requirement management outlines possible courses of action and patterns for describing a system's specifications and requirements. In complex biomedical systems development, a bioinformatics requirement change typically causes a ripple effect and forces the categorical requirements model to be altered as well.

MYCO-LIMS is used in the FungalWeb [14] integrated system to respond to queries regarding the clinical, pharmaceutical, industrial, and environmental processes related to pathogenic fungal enzymes and their related products. It is estimated that laboratory data account for 60–80% of the data generated during the entire clinical trial process [15].

The FungalWeb semantic Web infrastructure [16] (Figure 2) consists of the FungalWeb ontology, skin disease ontology (SKDON), a text mining framework, and intelligent agents. In addition, several external applications such as the MYCO-LIMS, the MYCO-LIS, and mutation miner [17] have been designed for knowledge exchange.

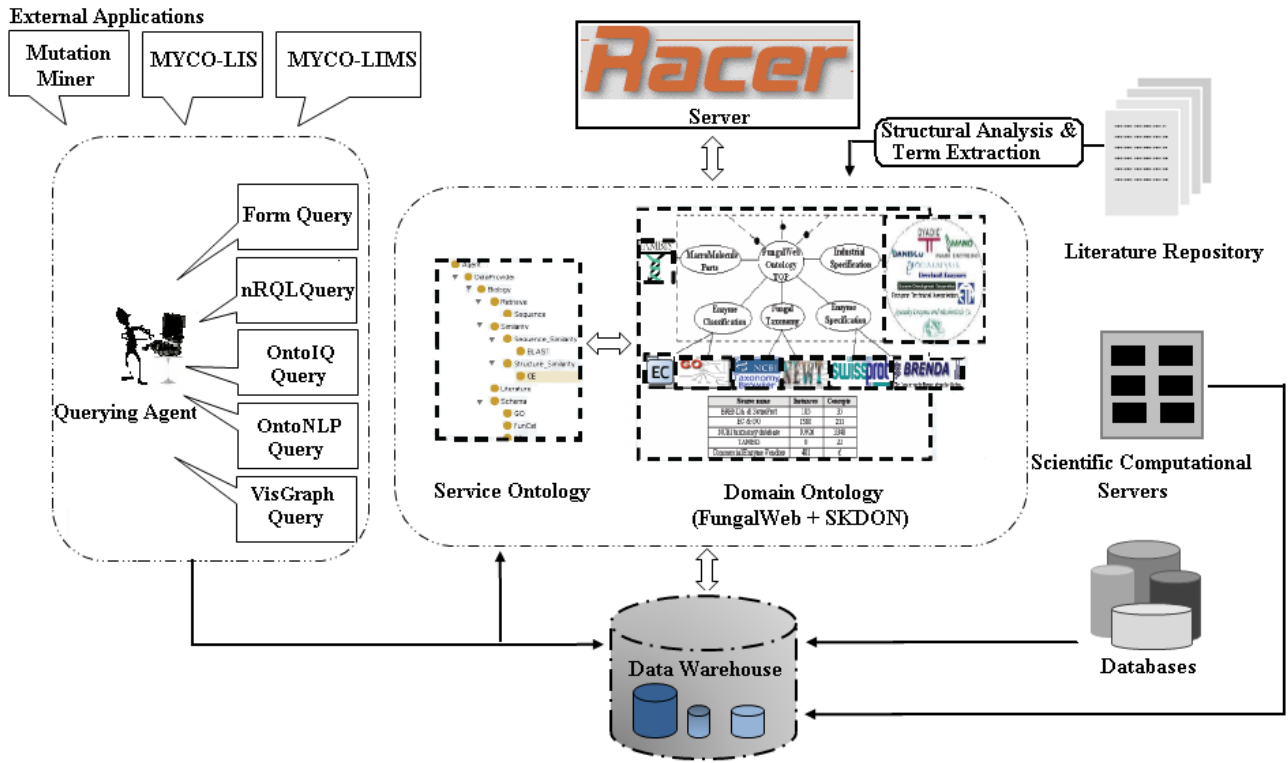

**Figure 2:** The FungalWeb infrastructure.

Microarrays are produced in different proportions, depending on the specific requirements of the gene expression study being initiated. A typical microarray may include thousands of distinct cDNA probes [18]. Preparation of an array begins with the clone set deliverance in the form of plates or tissue samples (with associated data) from a vendor or other source [18]. The MYCO-LIMS will be able to maintain the taxonomy for each plate or sample in the system; such that a user can easily see the life cycle of the entity. The LIMS is based on MGED-specified [19] microarray data exchange standards, such as MIAME [20] or MAGE-ML [21].

Software systems in general and MYCO-LIMS in particular are characterized both by their functional behaviour (what the system does) and by their nonfunctional behaviour (how the system behaves with respect to some observable attributes like reliability, reusability, maintainability, etc.). Both aspects are relevant to software development and are captured correspondingly as functional requirements (FRs) and nonfunctional requirements (NFRs).

## 2.1. LIMS Functional Requirements (FRs)

MYCO-LIMS is a Web-based system capable of providing services such as managing microarray gene expression data and laboratory supplies, managing patients, physicians, laboratories supplies or vendors' information, managing and tracking samples information, and managing orders. Figure 3 summarizes some of the main actors and services of the MYCO-LIMS application in a standard use case diagram.

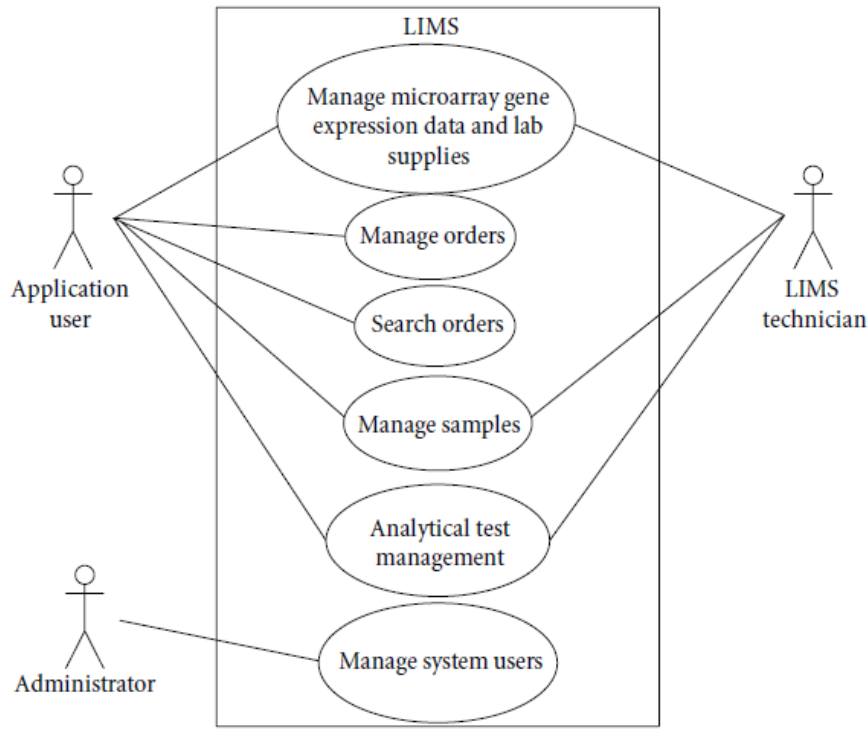

**Figure 3:** LIMS use case diagram.

MYCO-LIMS is capable of receiving multiple orders or cancelation requests at the same time. It requires its users to have a certain level of privileges to access any of the functionalities, except when searching for a product. The privileges are granted automatically upon successful authentication. In this paper, we limit the scope of the discussion to one functional requirement, "manage order", and further decompose it into two more specific sub-NFRs, "view orders," and "place order". In each decomposition, the offspring FRs contribute toward satisfying the goal of the parent. Figure 4 presents the functional model and shows that an FR is realized through the various phases of development by many functional models (e.g., in the object-oriented field, a use case model is used in the requirements engineering phase, a design model is used in the software design phase, etc.). Each model is an aggregation of one or more artifacts (e.g., a use case and sequences of events representing scenarios for the use case model, classes and methods for the design model). For instance, view order use case is refined to a sequence of events < enter order number, visualize order > illustrating an instance of viewOrder service; each event is refined as a method (viewOrder-Session.view and viewCatalogue.view correspondingly) in the design phase. Modeling FRs and their refinements in a hierarchical way gives us the option of decoupling the task of tracing FRs change from a specific development practice or paradigm. Figure 4 visualizes the FR hierarchical model for the chosen case study through the hierarchy graph that forms a primary taxonomy for analyzing ontological relationships between requirements.

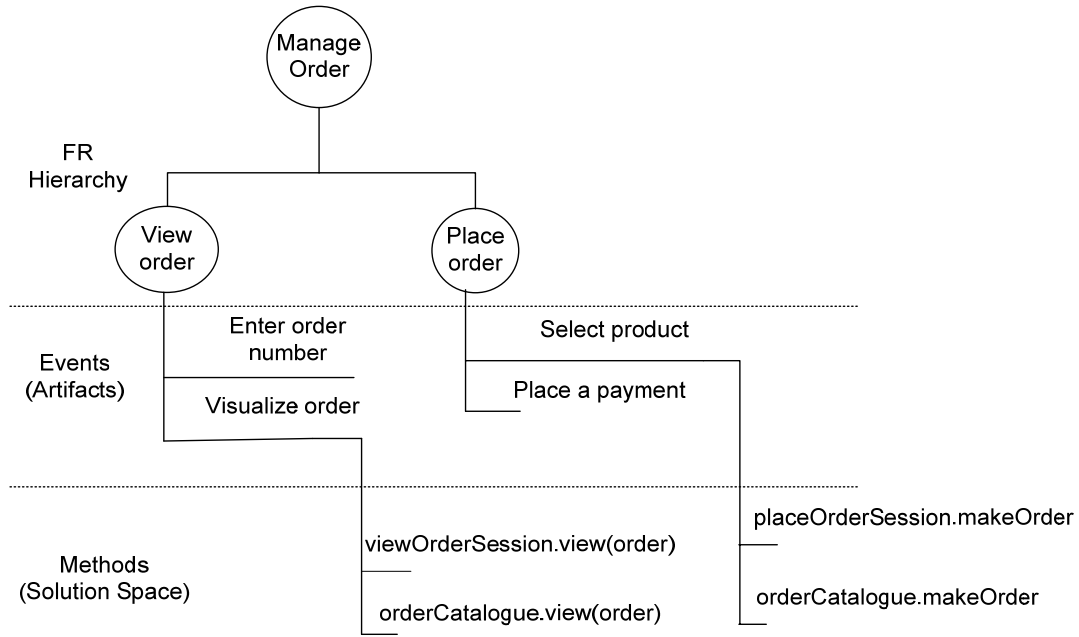

**Figure 4:** Illustration of the MYCO-LIMS FR traceability model.

## 2.2. LIMS Nonfunctional Requirements (NFRs)

The use case diagram shown in Figure 3 specifies the FRs of MYCOLIMS services. At the same time, compliance with the NFRs, such as performance, scalability, accuracy, robustness, accessibility, resilience, and usability, is one of the most important issues in the software engineering field today. NFRs impose restrictions by specifying external constraints on the software design and implementation process [22] and therefore need to be considered as an integral part of the process of conceptual modeling of the requirements. The goal of this section is to build a systematic, quantitative, and formal approach to NFR modeling, impact detection, and volatility evaluation/decision-making from the early stages of the software development process.

We decompose a high-level NFR into more specific sub-NFRs. In each decomposition, the offspring NFRs can contribute partially or fully toward satisficing the parent. Let us consider the requirements of "managing orders with good security" and "maintain the users' transactions with good performance". The security requirement constitutes quite a broad topic. To effectively deal with it, the NFR may need to be broken down into smaller components, so that an effective solution can be found. We can decompose the security NFR into the sub-NFRs integrity, confidentiality, and availability. In the security example, each sub-NFR has to be satisfied for the security NFR to be satisfied. The sub-NFRs are refined (operationalized) into solutions that will satisfice the NFR. These solutions provide operations, processes, data representations, structuring, constraints, and agents in the target system to meet the goals stated in the NFRs. In the confidentiality example, a solution can consist of either implementing authorization or the use of additional ID. Figure 5 visualizes the NFR partial hierarchy resulting from the decomposition and operationalization relations for the NFRs chosen in the LIMS.

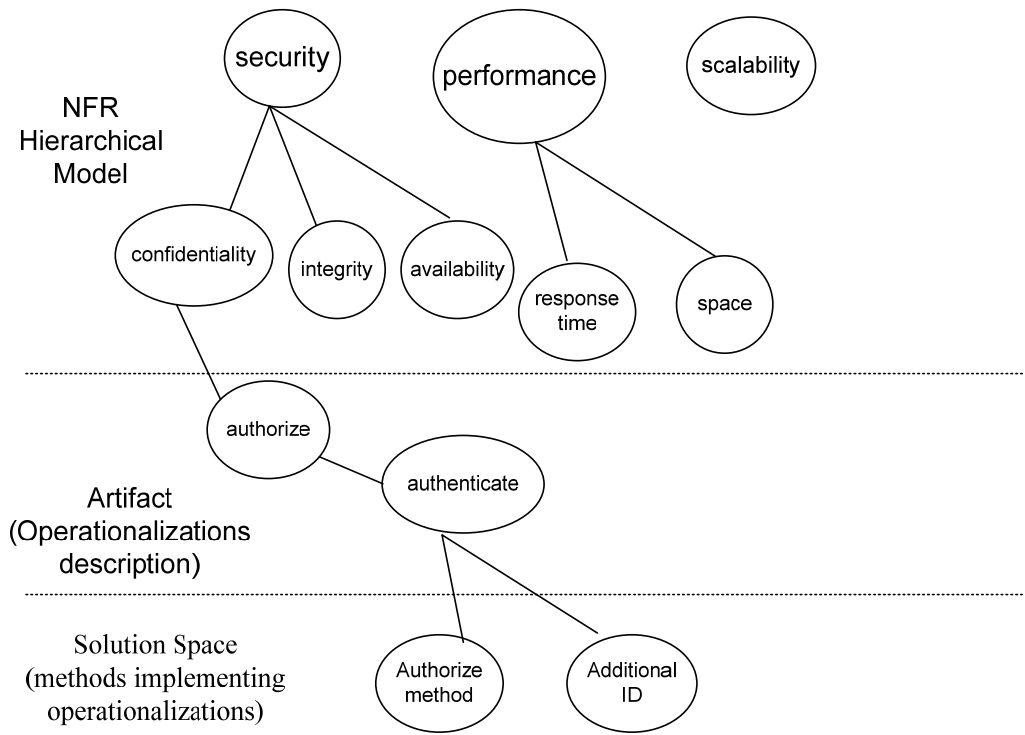

**Figure 5:** Illustration of the MYCO-LIMS NFR traceability model.

NFRs pose further challenges when it comes to determining their relationships with FRs. The tendency for NFRs to have a wide-ranging impact on a software system services and the strong interdependencies and tradeoffs that exist between them and the FRs leave typical existing software modeling methods incapable of integrating them into software engineering. In Section 2.3, we propose anew generic ontological framework for conceptualizing the NFR and FR requirements, their decompositions, and the corresponding associations.

## 2.3. Integrating FRs and NFRs into a generic ontological framework

Hardly any requirement is manifested in isolation, and normally the provision of one requirement may affect the level of provision of another. Understanding FR/NFR relations is essential to influencing the consistency and change management of the requirements. Once a software system has been deployed, it is typically straightforward to observe whether a certain FR has been met or not, as the ranges of success or failure in its context can be rigidly defined. However, the same is not true for NFRs as these can refer to quantitative statements that can be linked to other elements of the system. In fact, NFRs are not standalone goals as NFRs and their derived design solutions (operationalizations) can be associated to FRs throughout the software development process.

While tracing requirements is a major activity for change management of the system requirements, it has, by and large, been neglected for NFRs in practice. This area needs a special attention because NFRs are subjective in nature and have abroad impact on the system as a whole. In this section, we illustrate our approach toward finding an effective method for conceptualizing NFRs based on their hierarchy and their interrelations with FRs in the MYCO-LIMS invoicing system case study.

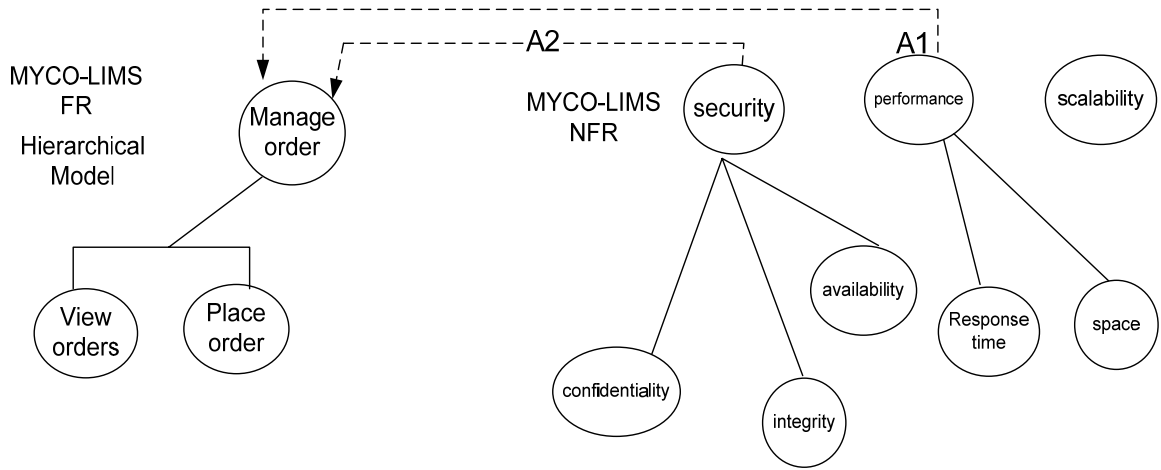

**Figure 6:** Illustration of MYCO-LIMS NFRs/FRs dependencies hierarchical model.

For example, associating response time NFR to view order use case would indicate that the software must execute the functionality within an acceptable duration (see association A1, Figure 6). Another example is associating security NFR to the "*manage order*" FR, which would indicate that the interaction between user and the software system in the "manage order" service must be secured (see association A2, Figure 6), which also precisely implies that user interface for other interactions is not required to be secured.

If an association exists between a parent NFR and a functionality (e.g. association *A2* between *security* and *manage_order*, or association *A1* between *performance* and *manage_order*) (see Figure 6), there will be an association between operationalizations derived from NFRs and methods derived from the functionality (e.g. *authorize* derived from *security*, and *placeOrderSession.makeOrder* derived from *manage_order*) (see Figure 7).

Figure 7 illustrates the refinement of the interactions. The complete change management model would require the refinement of performance and scalability into operationalizations and methods, and the identification of the associated interaction points to which they are mapped.

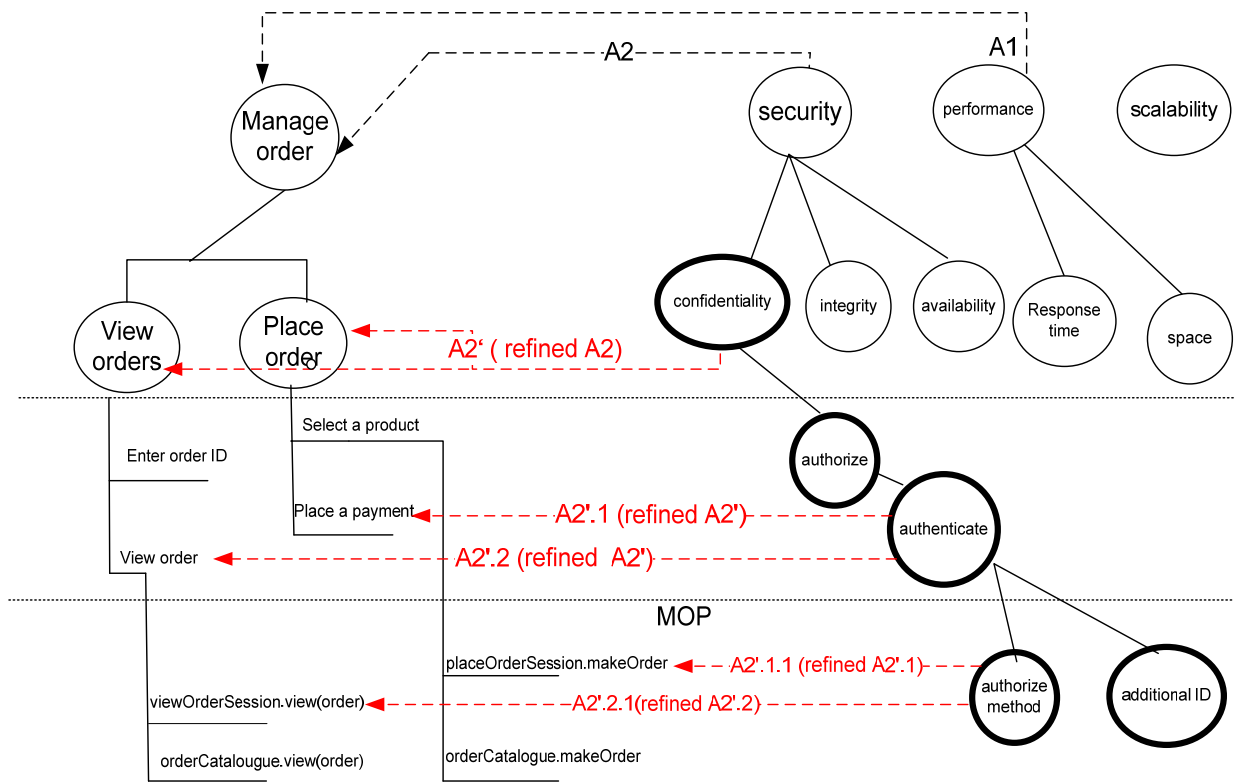

**Figure 7:** MYCO-LIMS Requirements associations' refinement.

A change in FRs or NFRs can be authorized if and only if that change is consistent with the existing requirements model. Our future work includes the development of more consistency rules based on a formal presentation of the FR and NFR hierarchies and their relations, and these rules will be checked automatically before a change is authorized.

The conceptualization of FR and NFR hierarchies and their interconnections form the bases for analyzing ontological relationships between requirements in the Service Ontology (see Figure 2). The NFR/FR ontological framework introduced in this section can be visualized through a categorical hierarchical graph, which makes it possible to keep track of the required behavior of the system using dynamic views of software behaviors from requirements elicitation to implementation.

The following subsection proposes a generic categorical model of requirements with an emphasis on NFRs and their interdependencies and refinements through using category theory as an advanced mathematical formalism, and this model will be is independent of any programming paradigm.

## 3. Generic Categorical Representation of Requirements and their Traceability

An ontology is a categorization of things in the real world. It can be viewed in terms of an interconnected hierarchy of theories as a sub-category of a category of theories expressed in a formal logic [23]. Categorical notations consist of diagrams with arrows. A category consists of a collection of objects and a collection of arrows (called morphisms). Each arrow $f: X \rightarrow Y$ represents a function. Representation of a category can be formalized using the notion of the diagram. We have chosen category theory as the main formalism in our framework because it has proved itself to be an efficient vehicle to examine the process of structural change in living and evolving systems [24].

In fact, we can use category theory to represent ontologies as a modular hierarchy of domain knowledge. Categories capture and compose the interactions between objects, identify the patterns of interacting objects in ontologies, and either extract invariants in their action or decompose a complex object in basic components. Categories are also able to identify patterns that recur again and again in a changing system. Other reasons for using category theory in our framework, as stated by Adamek et al. [25], are the abundance, precise language, and convenience of symbolism for visualization. Although category theory is a relatively new domain of mathematics, introduced and formulated in1945 [7], categories are frequently found in this field (sets, vector spaces, groups, and topological spaces all naturally give rise to categories). The use of categories can enable the recognition of certain regularities in distinguishing a variety of objects, their interactions can be captured and composed, equivalent interactions can be differentiated, patterns of interacting objects can be identified and some invariants in their action are extracted, and a complex object can be decomposed into its basic components [26].

In order to explicitly reason about the impact of NFRs and their refinements on the project throughout the software development process, we explicitly represent NFRs, FRs, and their dependencies and refinements using the language of category theory. Figure 8 captures the generic view on the requirements modeling process where requirements group, hierarchical model, artifacts, and solution space are categories representing the project requirements, the analysis models, the refined representations of the project requirements, and the requirements implementation, respectively. The arrows are morphisms which capture the refinement processes, namely, decomposition, operationalization, and implementation defined as shown in Figure 8.

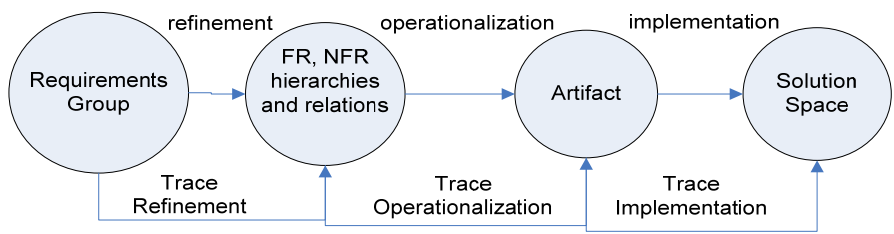

**Figure 8:** Generic categorical framework for requirement traceability.

Figure 8 shows that a requirement is realized through the various phases of refinement by hierarchical models, where each model is an aggregation of one or more artifacts. The implementation arrow refines the artifacts into solutions in the target system that will satisfy the requirements. These solutions provide operations, processes, data representations, structuring, constraints, and agents in the target system to meet the requirements represented in the requirements group. High-level FRs are refined in the requirements analysis phase into more specific sub-FRs (use cases and their relations (FR hierarchy model),e.g.), which are then operationalized as use case scenarios describing instances of interactions between the actors and the software, and modeled as events (artifacts), which are implemented as methods (solution space). High-level NFRs are refined into an NFR hierarchy where the offspring NFRs can contribute fully or partially toward satisficing the parent. The sub-NFRs are operationalized into solutions (artifacts) in the target systems, which will satisfice the NFR. These procedures provide operations, processes, data representations, structuring, constraints, and agents in the target system to meet the needs stated in the NFRs, and are implemented as methods in the solutions pace.

The requirement refinements are then expressed formally in terms of the composition operator °, assigning to each pair of arrows $f$ and $g$, with cod $f$ = dom $g$, a composite arrow $g \circ f$: dom $f \to$ cod $g$ (cod $f$ is a notation for a codomain, and dom $f$ is the notation used to indicate the domain of a function $f$). In this case, each requirement object belonging to the Requirements Group category will be refined to its implementation belonging to the Solution Space. The resulting solution forces preservation of the requirements and their relations, which are modeled with the *trace* arrows. The consistency between the solution and the original requirements can be guaranteed by the composition of categorical arrows representing morphisms. As a result, each change to a requirement or its refinement belonging to the domain of *f* will be traced to its refinement belonging to the codomain of *g* by means of the composition of the corresponding trace arrows.

**3.1. Categorical representation of FRs, NFRs hierarchies and their interdependencies**

The category *FR, NFR hierarchies, and relations* (Figure 9) consists of objects representing FRs and NFRs, their decomposition into sub-FR and sub-NFR (which are also FR and NFR correspondingly), and their impact associations; above concepts are treated jointly and in an integrated fashion. We identify four critical areas for impact detection in which NFRs require change management support: (i) impact of changes to FRs on NFRs (inter-model integration); (ii) impact of changes to NFRs on FRs (inter-model integration); (iii) impact of changes to NFRs on sub-NFRs and parent NFRs (intra-model integration); and (iv) impact of changes to NFRs on other interacting NFRs (intra-model integration).

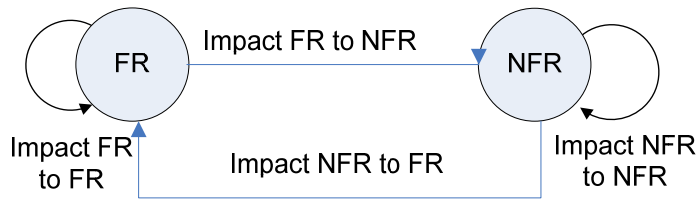

**Figure 9:** *FR, NFR hierarchies, and relations* category.

**3.2. Categorical representation of the Solution Space**

The Solution Space category contains State Space *SS* (all potential states including initial states), State Transition *ST* (next state function), Class *C* categorical objects, and Methods arrows. The *trace implementation* morphism traces the effect of the changes to Artifact objects on the Solution Space objects. In Figure 10, for instance, we illustrate the refinement of an event from the Artifact category to a state transition object *ST*. Moreover, each state transition *ST* is defined on the state space *SS* (arrow *ST_SS*) linked by a function *ST_C: ST →  C* to a class *C*. The state transitions are implemented by methods captured with the function *ST_M: ST → AP_M*, and belonging to a class *C* (see function *M_C*). The above functions support the tracing mechanism and are captured formally in Figure 10. The changes are then represented formally in terms of the composition operator °; for instance, $E\_ST \circ ST\_SS \circ ST\_C$ will trace a change in dom *E_ST* (which is *A_Event*) to the codomain of *ST_C* (which is Class *C*).

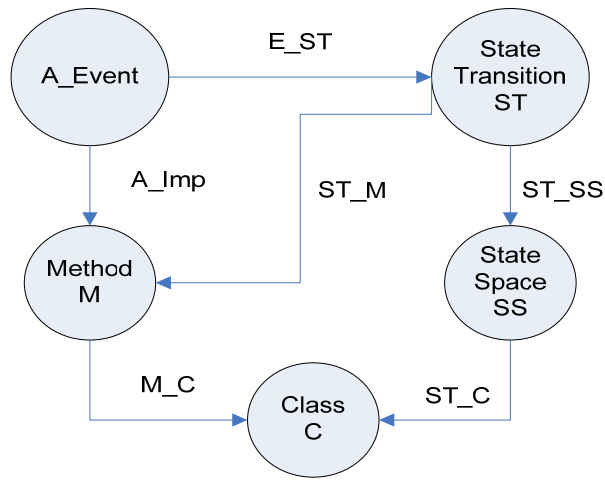

**Figure 10:** Tracing the changes to the state spaces, classes, and methods.

As we presented in [27], category theory has great potential as a mathematical vehicle to represent, track, and analyze changes in ontologies. For example, it can be used in the taxonomical representation of requirements to help in the study of the ontological relationship between the various nodes within the hierarchy. After describing the ontological concepts within the categories representing a modular hierarchy of domain knowledge, we have employed category theory to analyze ontological changes and agent interaction in different stages of the RLR framework [14].

## 4. The RLR Framework

The RLR multiagent framework [14] (RLR stands for: representation, legitimation, reproduction) (Figure 11) aimed at capturing, tracking, representing, and managing the changes in a formal and consistent way, enabling the system to generate reproducible results using change capture agents, reasoning agents, learning agents, and negation agents. Change capture agents are responsible for discovering, capturing, and tracking changes in ontology, by processing the change logs. The change logs accumulate important data about various types of changes. In RLR, a learner agent uses these historical records of changes that occur over and over in a change process to derive a pattern to estimate the rate and direction of future changes for a system by generating rules or models. The reasoner (which verifies the results of a change) and negotiation agents can change the rules generated and send modifications to the learning agent. Negotiation takes place when agents with conflicting interests want to cooperate. In RLR, the negotiation agent acts as a mediator allowing the ontology engineer and other autonomous agents to negotiate the best possible realization of a specific change, while maximizing the benefits and minimizing the loss caused by such a change. A human expert may then browse the results, propose actions, and decide whether to confirm, delete, or modify the proposals, in accordance with the intention of the application. In RLR, negotiation is defined based on the conceptual model of argumentation [28], where an argument is described as a piece of information allowing an agent to support and justify its negotiation stance or affect another agent's position through a communication language and a formal protocol [28]. The negotiation protocol can formally provide the necessary rules [29] (i.e., rules for admissions, withdrawals, and terminations) for negotiation dialog among participants. In our approach, we have partially adapted the architecture of the argumentative negotiating agent described at [30].

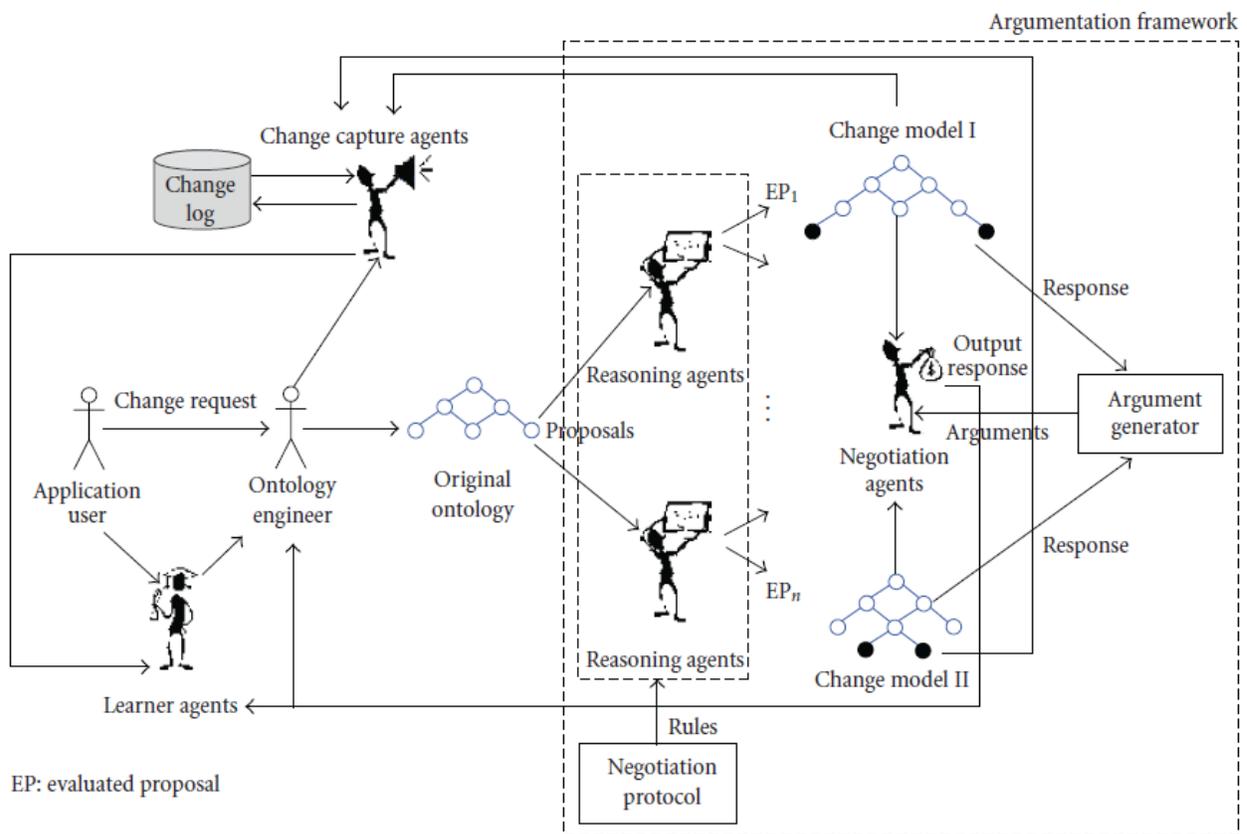

**Figure 11:** The RLR framework for change management and conflict resolution.

Within the RLR argumentative architecture, the negotiation agent and the reasoning agent provide arguments for the acceptance or rejection of a change proposal. The "argument generator" (Figure 11) determines appropriate responses based on the negotiation rules.

Different arguments attack one another to impose their rules and defeat their peers by sending counter arguments. The inferred arguments can increase the possibility of higher-quality agreement [30, 31].The negotiation protocols in the RLR architecture contain the negotiation protocol's rules, which dictate a protocol. As an application is used and evolves over time, the change logs accumulate invaluable data and information about various types of changes. A learner agent can use these historical records of changes that occur over and over in a change process to derive a pattern out of the rules generated. The reasoner and the negotiation agents can change the rules—if necessary—and send modifications to the learning agent. The learning agent starts with limited, uncertain knowledge of the domain and tries to improve itself, relying on adaptive learning based on the semantics provided by the ontological backbone.

## 5. Employing Category Theory in the RLR Framework

We have used categories in various stages of the RLR multiagent framework for representing and tracking changes in NFRs and FRs.

### 5.1. Category theory for representing and tracking changes

The categorical representation enables the progressive analysis of ontologies and can be used to represent the evolutionary structure of an ontology, to provide facilities for tracking each change and to analyze the impact of these changes by the following:

   a) **Comparing different states of a class:** We have used "functor", which is a morphism in the category of all small categories (where classes are defined as categories) to describe the set of state space (set of all possible states for a given state variable set) for a class as a cross product of attribute domains and the operations of a class as transitions between states for ontological elements indexed by time. Using the functor, the transition from $O_t$ to $O_{t'}$, where the time changes from $t$ to $t'$, can be represented and analyzed. For more information see [27].

b) **Measuring coupling:** Coupling indicates the complexity of evolving structure [27]. When coupling is high, it indicates existence of a large number of dependencies in an ontological structure which must be checked to analyze and control the chain of changes. Following [32], to analyze the coupling we consider three types of arrows namely: pre-condition, post-condition and message-send arrows in category theory to analyze various conditional changes [27].

c) **Using Pushout and Pullback:** When a change is either integration or mergence, one can use two categorical constructors: pushout and pullback [33]. The pushout for two morphisms $f: A \rightarrow B$ and $g: A \rightarrow C$ is an object $D$, and two morphisms $i_1: B \rightarrow D$ and $i_2: C \rightarrow D$, such that the square commutes (Figure 12(a)). $D$ is initial object in the full subcategory of all such candidates $D'$ (i.e. for all objects $D'$ with morphisms $j_1$ and $j_2$, there is a unique morphism from $D$ to $D'$).

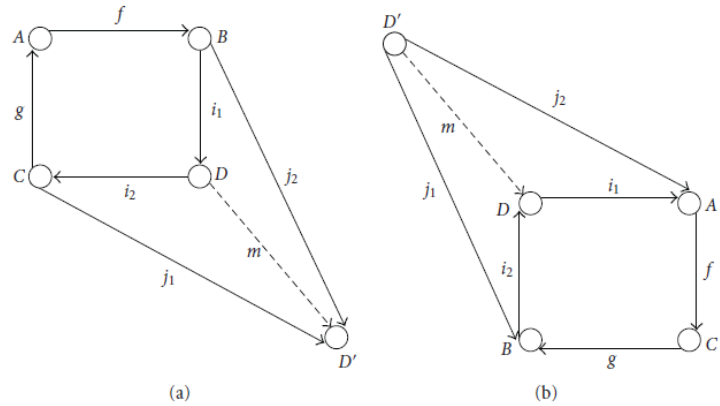

**Figure 12:** Two categorical constructors: (a) Pushout, (b) Pullback.

The *pullback* (also known as "Cartesian square") for two morphisms *f: A→C* and *g: B→C* is an object *D*, and two morphisms $i_1$: *D→A* and $i_2$: *D→B*, such that the square commutes. Here *D* is the terminal object in the full subcategory of all such candidates *D´* [34] (Figure 12(b)). Hitzler et al. [35] and Zimmermann et al. [36] also used pushout for ontology alignment.

**5.2. Category Theory for Representing Agent interactions and Conflict Resolution**

Intelligent agents perform actions in a context by using rules. Changing the rules is a main adaptation principle [37] for learning in RLR framework. The adaptive agents in the RLR have been defined following Resconi's method [37]. The rules consist of a set of semantic unity symbolized by $S_1$, IN, $P_1$, and OUT, representing the input statement, the domain of the rule, the rule, and the range of the rule (denoting the value of an agent's action), respectively. When we are working in a dynamic environment, it is likely that these rules change into other rules. Therefore, a single change in the primary structure triggers other changes in rules and contexts. A communication channel [37] between those rules and between different adaptive agents is needed to manage all the necessary interactions.

In the RLR we have used category theory formalism, along with general systems logical theory (GSLT) [38], to formalize agents' communications. For instance, the communication between different semantic unities [14] can be represented as in Figure 13.

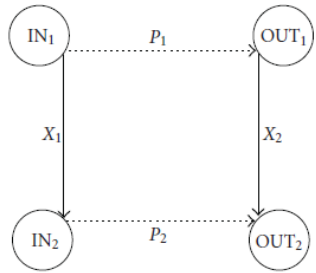

**Figure 13:** The categorical representation that shows how rules $P_1$ and $P_2$ enable the transformation of rule $X_1$ into rule $X_2$ (following [37]).

In addition, category theory can be used for modeling agent interactions [39], yielding a practical image of adaptive learning agents, their semantic unities, and adaptation channels [37].

We have also followed the approach presented in [40] for representing the product and coproduct of objects, to categorically represent the integration and merging of NFR objects, which are defined as ontological elements. The negotiation agent in RLR can negotiate to determine the best of several methods of integration. For example, an integration can be implemented as the product A×B (all possible pairs < elements from A, elements from B >), or the coproduct of the objects A+B (all elements from A and all elements from B) for both categorical objects and arrows (denoting ontological elements). Assume that we define the following arguments for integrating ontological structures within a dialectical database [31] in the RLR framework:

$$a_1: A \times B, \quad a_2: A+B, \quad a_3: A, \quad a_4: B$$

Categorically speaking "$a_1$ defeats $a_2$" can be represented by an arrow from $a_1$ (domain) to $a_2$ (codomain) (Figure 14). By following categorical representation, an argumentation network will be generated, which can be used to formally describe negotiations and speed up inferences [31].

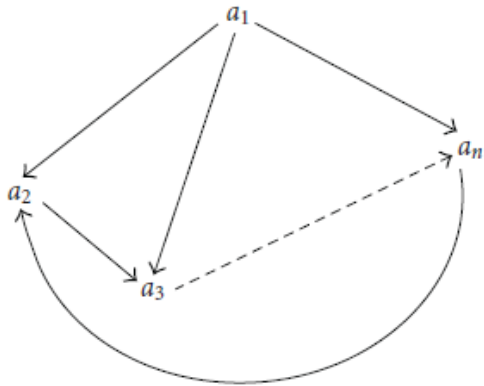

**Figure 14:** Categorical representation of the argumentation network.

## 6. Application Scenarios

As shown in section 5, category theory can be used in RLR to integrate time factor and represent and track changes in ontological structure in time through using the notion of state capturing an instance of system's FRs, NFRs and associations at certain period of time. For example, a change in the Authorize Method would affect the method "placeOrderSession.makeOrder" in state $St_1$ of the system, which will be traced to changes in state $St_2$ (Figure 15). Explicitly capturing of the evolution of the requirements in time can aid

MYCO-LIMS developers and maintainers to deal with requirements change management in highly dynamic clinical applications.

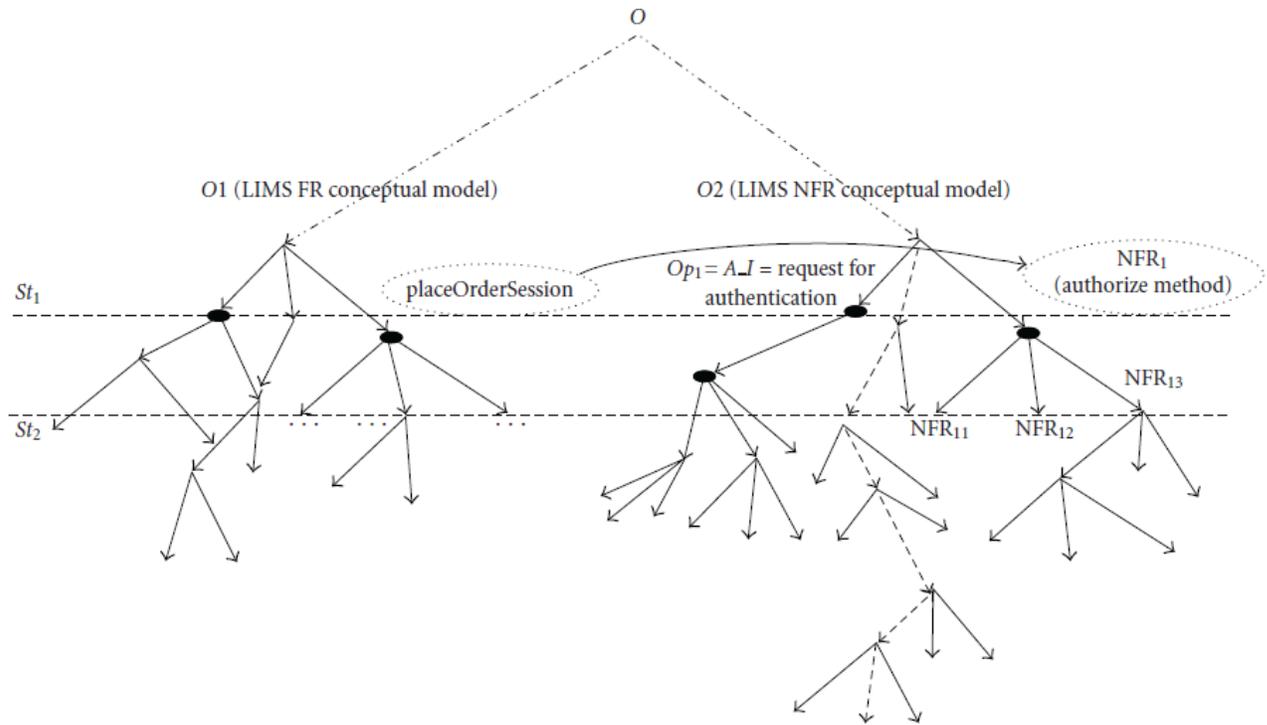

**Figure 15:** Categorical representation of evolving MYCO-LIMS functional requirements (FRs) and nonfunctional requirements (NFRs).

Generally speaking, changes to each NFR would lead to changes in the conceptual framework. As mentioned in Section 3, we are monitoring the effect of FR or NFR changes through their refinement relations, that is (1) identifying the "slice" of the conceptual framework that will be affected by the change, (2) applying the consistency rules to make sure the change does not introduce any inconsistencies in the "slice", and (3) implementing the change, if authorized.

The RLR change management framework is modeled as an intelligent control loop, which has one state for each of the above stages (1), (2), and (3), the events modeling the change of state. Considering the requirements to be organized in a lattice-like ontological framework, in order to represent the various states of our conceptualization, we use a categorical discrete state model, which describes the states and events in the ontological structure using a diagrammatical notion. The discrete state model is specified by a state space (all potential states), a set of initial states, and a next-state function. Based on our application, we designed our class diagrams following the method described in [27, 32] (Figure 16), which can be used to create patterns for learning agents. The $Op_i$ arrows in this figure represent the operations for the class, where in the operation or event $Op_1$ causes an object in state $St_1$ to undergo a transition to state $St_2$. The operation $Op_1$ has no effect on the object if it is in any other state, since no arrow labeled $Op_1$ originates from any other state. The object ∅ the diagram is the null state. The create arrow represents the creation of the object by assigning an identifier and setting its state to the initial defined state, and the destroy arrow represents its destruction [32].

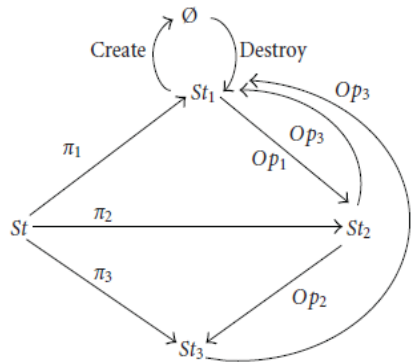

**Figure 16:** Class diagram for the part of the FR-NFR ontological structure that represents the transition between states.

Based on [32], a projection arrow for any attribute is drawn from the state space to the attribute domain and labeled with the name of the attribute (i.e. $\pi_i$ represents the value of the

*i*th attribute). A selection arrow for each state *x* (labeled as σ$_x$) is drawn from the state space to that state (i.e. σ$_i$ gives the *i*th state).

Using category theory we represent the most common operations during requirement change management such as adding/deleting a class of requirements, combining two classes of requirements into one, adding a generalization/association relationship, adding/deleting a property or relationship. For more information see [27].

**7. Evaluation of the Approach along with Change Verification**

The legitimation phase in RLR verifies the legitimacy and consistency of a change in the domain of interest. This phase assesses the impact of a potential change before the change is actually made. Experts and logical reasoners study a change based on its consistency with the whole design, in varying degrees of granularity. Then, final approval is needed from the end users. Logical legitimation is obtained by a reasoning agent, which is a software agent that controls and verifies the logical validity of a system, revealing inconsistencies, misclassifications, hidden dependencies, and redundancies. It automatically notifies users or other agents when new information about the system becomes available. We use RACER [41] as a description logic reasoner agent, along with other semiformal reasoners in RLR. When the agent is faced with a change, it ought to revise its conceptualization based on the new input by reasoning about the consistency of the change using both prior and new knowledge. We also use a semi-automated reasoning system for basic category theory reasoning [42] based on a first-order sequent calculus [43], which captures the basic categorical constructors, functors, and natural transformations, and provides services to check consistency, semantic coherency, and inferencing [43]. Placing a new class of requirements in a system may sometimes lead to redundancy in the requirement taxonomy. One of the major issues in requirement analysis is finding and identifying logically equivalent classes and relationships which may differ in name but perform the same function. Employing category theory enables us to deal with this problem of logical equality in the evolving requirement hierarchy using isomorphic reasoning [44].

## 8. Related Work

Several efforts have been reported [45–48] during the last decade in the pursuit of inclusive frameworks for managing dynamic taxonomies, ontologies, and control vocabularies. Since existing knowledge representation languages, including well-established description logic, cannot guarantee the computability of highly expressive time-dependent models, the current efforts have been entirely focused on time-independent ontological models. However, the real ontological structures exist in time and space. From another perspective, those who choose other knowledge representation formalisms, such as state machine [49], can cope with time-based models, but these formalisms fail to address ontological concepts and rules because they are much too abstract and have no internal structure or clear semantics. In our proposed framework, category theory, with its rich set of constructors, can be considered as a complementary knowledge representation language for capturing and representing the full semantics of evolving abstract requirements conceptualized within ontological structures. Rosen [50] was among the first to propose the use of category theory in biology, in the context of a "relational biology".

Category theory also has been used by MacFarlane [24] as an efficient vehicle to examine the process of structural change in living/evolving systems. Whitmire [32], Wiels and Easterbrook [51], and Mens [52] have examined category theory for change management in software engineering domain. Hitzler et al. [35] and Zimmermann et al. [36] also have proposed using this formalism in knowledge representation area.

## 9. Discussion, Challenges, and Future Work

Any attempt to successfully systematize and automate electronic communication in biomedicine — with its continuously changing nomenclature and requirements — needs to pay special attention to managing requirement volatility in various stages of the biomedical application life cycle. Due to the wide variety of requirements controlled by the LIMS across diverse industries, LIMS software needs to be inherently more flexible [15]. One of the issues in requirement evolution and change management is a lack of formal change models with clear and comprehensible semantics. In order to represent, track, and manage requirement changes throughout a LIMS software project, we have proposed an agent-based framework to handle evolving requirements, which are categorized in an ontological structure. An ontology provides a means for formally capturing the FR and NFR hierarchies and their interrelations, and for exhaustive tracing of the effects of a change on the conceptual framework. In addition, we have proposed using category theory—which is an intuitive and powerful formalism, independent of any choice of ontology language—to capture the full semantics of evolving hierarchies in various phases of RLR. It also provides a language to precisely describe many similar phenomena that occur in different mathematical fields with an appropriate degree of generality. For example, category theory makes it possible to make a precise distinction between categories via the notion of natural isomorphism. It also provides a unified language to describe topological spaces via the notion of concrete isomorphism [25]. In addition, categorists have developed a symbolism for visualizing complicated facts by means of diagrams. Our proposed method for employing category theory to manage the evolving FR and NFR hierarchical structure can significantly help formalize agile requirement modeling in highly dynamic clinical applications. Moreover, this method can be easily adapted to different project situations and needs. The ontology-grounded categorical framework introduced here can be used to reduce requirement volatility by facilitating the definition of consistency rules for requirement change and supporting the automatic evaluation of consistency rule compliance with software requirements. The knowledge captured about requirement volatility and formalized using category theory is a suitable means to trace the effect of any requirement change on the specifications of the whole system.

In the process of employing category theory as the core formalism for our proposed framework, we had to deal with several challenges. Some of the major ones included the reasoning issues and managing conceptualization changes. Although we are able to provide some sort of basic reasoning and inferencing for categories, we still need to improve the reasoning capability to cover more advanced reasoning services. Also, the representation of changes in conceptualization due to the nature of NFRs, which needs to deal with abstract concepts and notions, is challenging. In order to overcome this issue, we are working on grammatical change algorithms in linguistics and language evolution. For future work, we plan to concentrate on the evolution of requirement calculation rules, which are based on the available requirement traceability information. Finally, a third field of study will address dashboard visualization and customization for various FungalWeb requirement management tools.